\documentclass[sigconf,nonacm]{acmart}
\usepackage[shortlabels]{enumitem}
\usepackage{xcolor}
\usepackage{pgfplots}
\pgfplotsset{compat=1.18}
\usepackage{booktabs}
\def\BibTeX{{\rm B\kern-.05em{\sc i\kern-.025em b}\kern-.08emT\kern-.1667em\lower.7ex\hbox{E}\kern-.125emX}}
    
\begin{document}
%
% The "title" command has an optional parameter, allowing the author to define a "short title" to be used in page headers.
\title{Agentic Multi-Source Grounding for Enhanced Query Intent Understanding: A DoorDash Case Study}

% used to denote shared contribution to the research.

\author{Emmanuel Aboah Boateng}
\email{e.aboahboateng@doordash.com}
\affiliation{%
  \institution{DoorDash, Inc.}
  \city{San Francisco}
  \state{California}
  \country{USA}
}

\author{Kyle Macdonald}
\email{kyle.macdonald@doordash.com}
\affiliation{%
  \institution{DoorDash, Inc.}
  \city{San Francisco}
  \state{California}
  \country{USA}
}
\author{Akshad Viswanathan}
\email{akshad.viswanathan@doordash.com}
\affiliation{%
  \institution{DoorDash, Inc.}
  \city{San Francisco}
  \state{California}
  \country{USA}
}

\author{Sudeep Das}
\email{sudeep.das2@doordash.com}
\orcid{0000-0002-1754-5811}
\affiliation{%
  \institution{DoorDash, Inc.}
  \city{San Francisco}
  \state{California}
  \country{USA}
}

\begin{abstract} 
  Accurately mapping user queries to business categories is a fundamental Information Retrieval challenge for multi-category marketplaces, where context-sparse queries such as ``Wildflower'' exhibit \emph{intent ambiguity}, simultaneously denoting a restaurant chain, a retail product, and a floral item. Traditional classifiers force a winner-takes-all assignment, while general-purpose LLMs hallucinate unavailable inventory. We introduce an Agentic Multi-Source Grounded system that addresses both failure modes by grounding LLM inference in (i)~a staged catalog entity retrieval pipeline and (ii)~an agentic web-search tool invoked autonomously for cold-start queries. Rather than predicting a single label, the model emits an ordered multi-intent set, resolved by a configurable disambiguation layer that applies deterministic business policies and is designed for extensibility to personalization signals. 
  This decoupled design generalizes across domains, allowing any marketplace to supply its own grounding sources and resolution rules without modifying the core architecture. Evaluated on DoorDash's multi-vertical search platform, the system achieves +10.9pp over the ungrounded LLM baseline and +4.6pp over the legacy production system. On long-tail queries, incremental ablations attribute +8.3pp to catalog grounding, +3.2pp to agentic web search grounding, and +1.5pp to dual-intent disambiguation, yielding 90.7\% accuracy (+13.0pp over baseline). The system is deployed in production, serving over 95\% of daily search impressions, and establishes a generalizable paradigm for applications requiring foundation models grounded in proprietary context and real-time web knowledge to resolve ambiguous, context-sparse decision problems at scale.
  \end{abstract}

% include context on the generalizability of this work in the abstract: 
% We developed this in-house and it's generalizable to several other use cases 
% peerhpas the abstract might be the best place to put this 
% The pluggable context is not clear: So perhaps the configurable would be the right phrasing 
% 


\maketitle

\section{Introduction}
\label{sec: intro}
\begin{figure}[t!]
  \centering
  \includegraphics[width=\columnwidth]{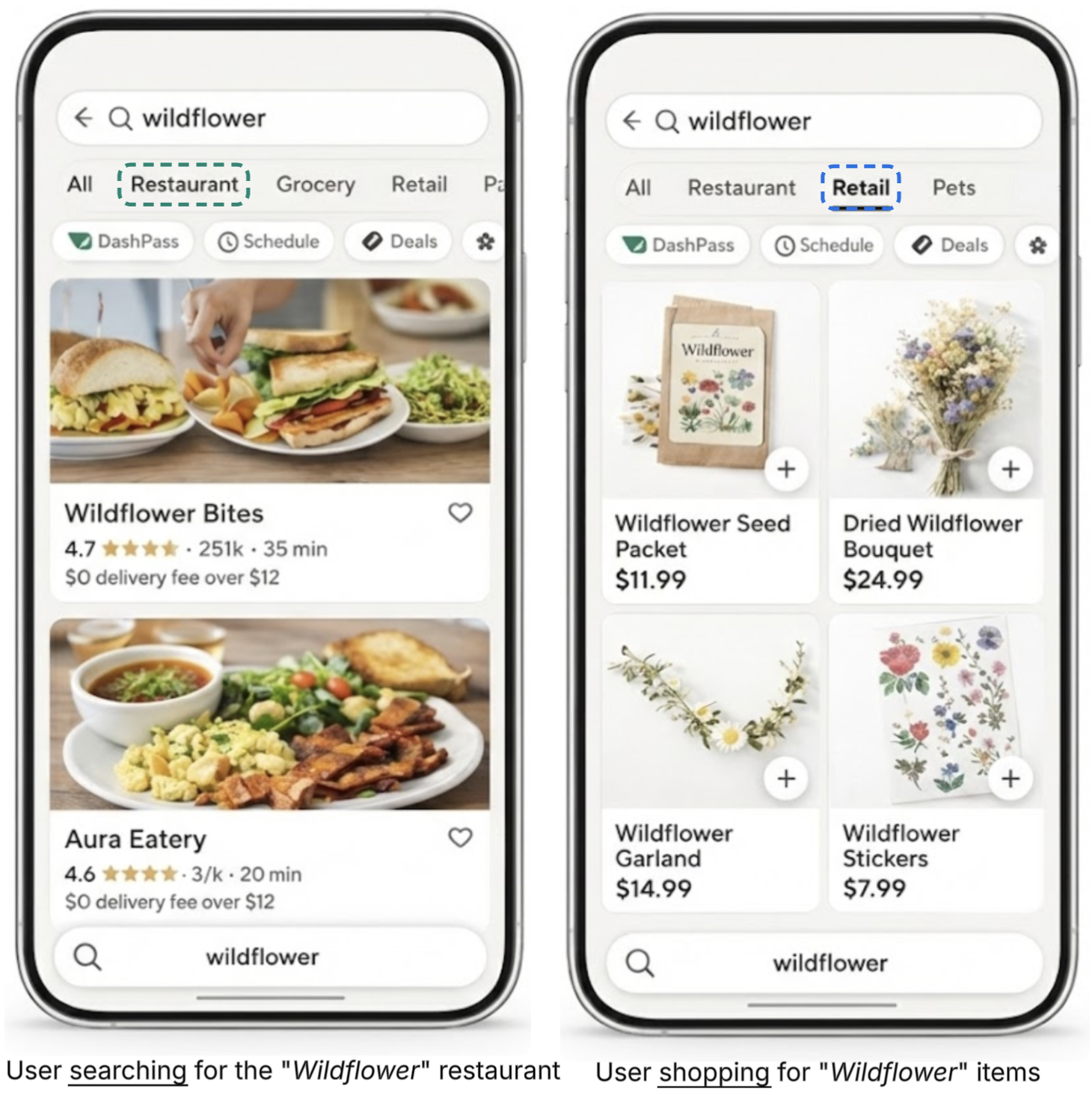}
    \caption{Intent Ambiguity for the query ``Wildflower.'' (Left) The Restaurant category surfaces the fictional Wildflower Bites chain. (Right) The Retail category surfaces literal flower products. Our system captures both interpretations to enable context-aware routing.}
    \Description{Side-by-side search results for the query Wildflower. The left side shows a restaurant interpretation represented by the fictional Wildflower Bites chain, and the right side shows retail flower products, illustrating why the query has multiple valid intents.}
  \label{fig:wildflower_ambiguity}
\end{figure}

The evolution of on-demand delivery platforms from single-category marketplaces (e.g., food delivery only) to multi-category ecosystems encompassing Food, Grocery, Retail, and Alcohol \cite{das2024applications} has introduced new challenges for search and discovery. As discussed in the aggregated search \cite{lalmas2011aggregated, diaz2009integration, arguello2009sources} and intent taxonomy \cite{broder2002taxonomy} literature, presenting results from multiple business categories on a single results page is challenging, especially when short, context-sparse queries can plausibly map to several categories simultaneously \cite{yuan-etal-2025-semi}.

Query Intent Understanding (QIU) predicts a user's shopping intent from their search query to route retrieval and ranking to the most relevant inventory. In a multi-category marketplace, this prediction directly influences which category experiences (e.g., Restaurant, Grocery, Retail) are surfaced; misrouting at this stage dominates downstream relevance regardless of ranking quality. Consider the query ``Wildflower'' (Figure~\ref{fig:wildflower_ambiguity}). A literal interpretation implies a Flower Product intent. However, ``Wildflower'' could also be a popular restaurant chain (e.g., the fictional ``Wildflower Bites'').\footnote{For trademark compliance, merchant names and associated imagery in examples throughout this paper are fictionalized representations of real ambiguity patterns observed in production.} If the system biases towards the literal meaning when the user intends to order lunch, the customer is presented with bouquets instead of the restaurant they expect, leading to a broken journey.

This problem has two distinct failure modes that existing approaches struggle to address simultaneously:

\textbf{Challenge 1}: Winner-Takes-All Classification.
Traditional supervised multi-class classifiers \cite{tsoumakas2007multi} force a single-label assignment. While such models can emit top-$k$ labels at inference, the softmax training objective encourages competition for probability mass, suppressing concurrently plausible intents under ambiguity. Increasing confidence in ``Restaurant'' necessarily suppresses ``Retail,'' discarding the valid alternative. This paradigm is fundamentally ill-suited for queries that span multiple categories \cite{yin2025midlm, zhou2025single}.

\textbf{Challenge 2}: LLM Hallucination Without Grounding.
To address the data sparsity of long-tail queries, recent work has turned to Large Language Models for their reasoning capabilities \cite{brown2020gpt3, luo2024large, ai2023query, zhang2025reic, srinivasan-etal-2022-quill}. However, generic LLMs frequently hallucinate availability \cite{ji2023survey, joren2025sufficient} without platform-specific grounding. An ungrounded model might misclassify ``Wildflower'' as a floral product, missing the popular restaurant chain entirely, or assume ``450 North'' is a restaurant when it is in fact a craft brewery. Without evidence from the platform's own catalog, the model's reasoning is disconnected from the marketplace's actual inventory.

While retrieval-augmented generation \cite{lewis2020rag} and tool-augmented LLMs \cite{schick2023toolformer} have been studied independently, to our knowledge no prior work combines agentic multi-source grounding with dual-intent prediction and pluggable disambiguation for production query intent understanding. We address both failure modes with two novel, complementary contributions:

\begin{itemize}
  \item \textbf{Agentic Multi-Source Grounded Classification} (Figure~\ref{fig:architecture}, steps 2--6). We ground LLM intent inference in platform-specific evidence by injecting high-precision catalog entities, retrieved via a staged semantic and fuzzy retrieval pipeline \cite{lewis2020rag, gao2023retrieval, srinivasan-etal-2022-quill}, directly into the LLM prompt. For cold-start queries absent from the catalog, the model autonomously invokes an agentic web-search tool to retrieve real-time world knowledge. This multi-source approach bridges the gap between static inventory data and emerging consumer trends.
  \item \textbf{Dual-Intent Prediction with Configurable Disambiguation} (Figure~\ref{fig:architecture}, steps 5--8). Rather than forcing a single label \cite{yin2025midlm}, we predict an ordered intent set $S = (v_p, v_s)$ and resolve it through a modular disambiguation layer. This layer currently uses deterministic pairwise business rules derived from historical win rates, but its configurable design supports extensibility to personalization signals such as user location or order history without architectural changes. This decoupling of prediction from resolution generalizes beyond DoorDash: any multi-category marketplace (e.g., an e-commerce platform distinguishing between brand stores and product listings) can supply its own grounding sources and disambiguation policies.
\end{itemize}

We validate our approach on DoorDash's multi-category search platform across four benchmarks. The system achieves +10.9pp accuracy over the ungrounded LLM baseline and +4.6pp over the legacy hybrid BERT+LLM production system, while maintaining high accuracy on high-frequency head queries. The system is currently deployed in production, serving over 95\% of daily search impressions via an offline batch-and-cache architecture.

\section{System Architecture}
\label{sec: 3method}
\begin{figure*}[t]
  \centering
  \includegraphics[width=0.85\textwidth]{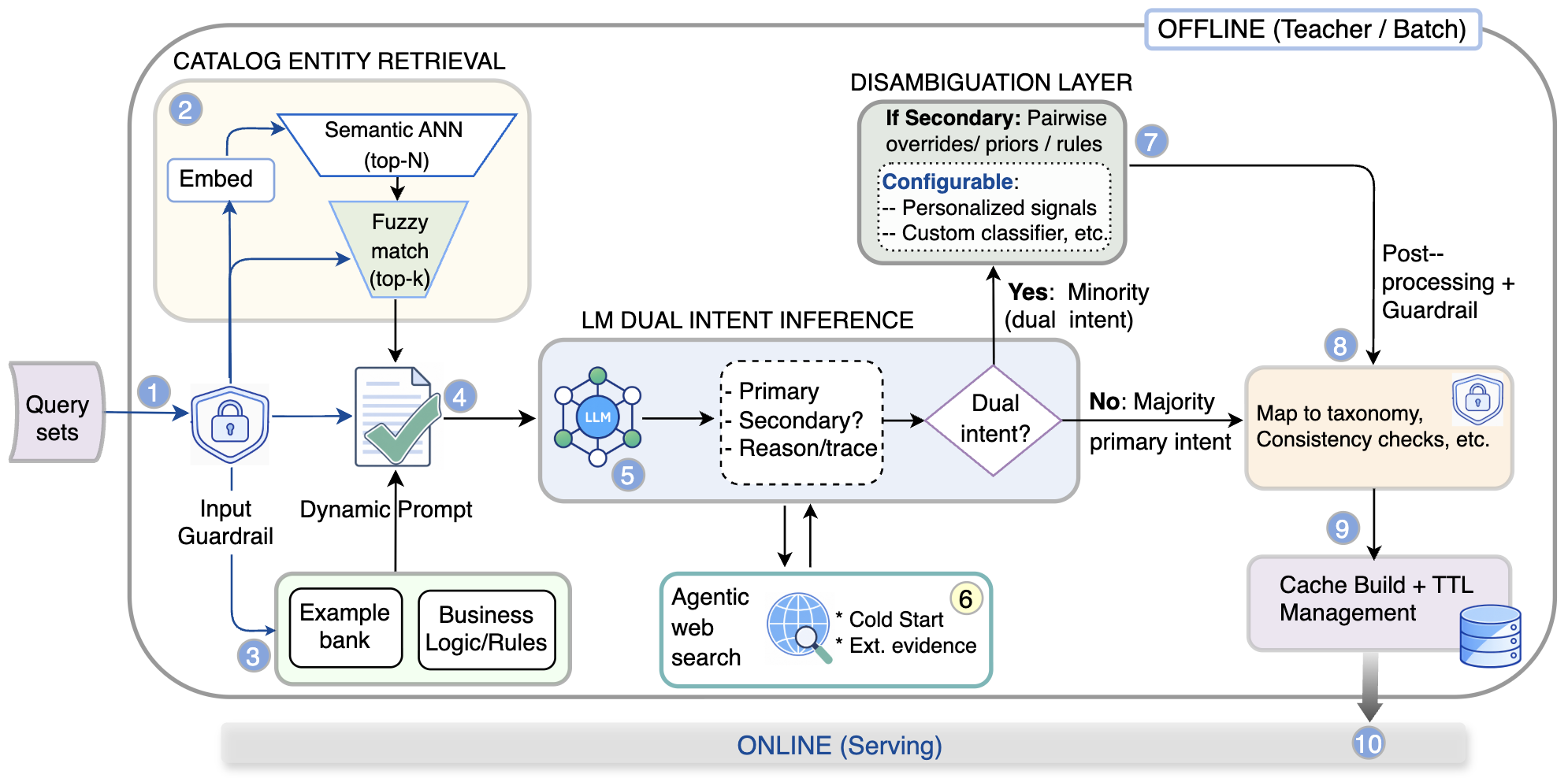}
  \caption{System Architecture Overview. The pipeline illustrates the multi-source evidence retrieval process (steps 2--4, 6), the dual-intent reasoning engine (step 5), and the pluggable disambiguation layer (step 7) that populates the production cache.}
  \Description{Architecture diagram for the query intent system. A user query flows through catalog retrieval and optional external web search, both evidence sources are injected into an LLM that predicts primary and secondary intents, and a downstream disambiguation layer resolves the final category before writing results into a production cache.}
  \label{fig:architecture}
\end{figure*}

Our solution moves away from pure text classification to a Retrieval-Augmented Classification approach \cite{lewis2020rag, srinivasan-etal-2022-quill}, where an LLM reasoning engine is grounded in multi-source evidence before emitting intent predictions (Figure~\ref{fig:architecture}).

\subsection{Formulation}
Given a query $q$ and a set of business categories $V = \{v_1, v_2, \ldots, v_n\}$, the conventional approach seeks the single category $v^*$ that maximizes the conditional probability:
\begin{equation}
    v^* = \operatorname*{argmax}_{v \in V} P(v \mid q)
\end{equation}
This formulation discards plausible alternatives. Drawing on vertical selection and federated search formulations \cite{arguello2010vertical, shokouhi2011federated}, we instead frame intent prediction as a set coverage problem. We optimize for the smallest set $S \subseteq V$ such that the probability of the user's true intent $v^*$ being contained in $S$ is maximized, conditioned on both the query and multi-source evidence $E$:
\begin{equation}
    S = \operatorname*{argmax}_{S \subseteq V,\; |S| \le k} P(v^* \in S \mid q, E)
\end{equation}
We constrain $|S| \le k{=}2$ as a pragmatic design choice that captures a salient alternative while keeping downstream disambiguation tractable. The evidence $E = E_{cat} \cup E_{ext}$ is the union of catalog-retrieved entities $E_{cat}$ and external search signals $E_{ext}$, detailed below.

\subsection{Multi-Source Evidence Retrieval}
To mitigate hallucinations and ground predictions in platform-specific knowledge, we inject evidence into the LLM prompt from two complementary sources.

\subsubsection{Staged Catalog Entity Retrieval}
We employ a two-stage retrieval architecture (Figure~\ref{fig:architecture}, steps 2--3) that balances recall and precision.

\textbf{Stage 1}: Semantic Retrieval. We retrieve a candidate set $C_{sem}$ using a dense retrieval index. Query $q$ and catalog entities $e \in \mathcal{E}$ are mapped to a shared embedding space via an encoder model $\phi$, and the top-$N$ candidates are retrieved using Approximate Nearest Neighbor (ANN) search:
\begin{equation}
    C_{sem} = \operatorname*{top\text{-}N}_{e \in \mathcal{E}} \; \text{cos}\big(\phi(q),\; \phi(e)\big)
\end{equation}
This ANN approach enables low-latency retrieval at scale, capturing conceptual matches even under severe lexical drift (e.g., ``better chew'' $\rightarrow$ Better Chew plant-based brand).

\textbf{Stage 2}: Fuzzy Refinement. To filter low-precision semantic matches and handle lexical variations such as typos, we re-rank $C_{sem}$ using a weighted fuzzy matching score. For a query $q$ and candidate entity $e$:
\begin{equation}
    S_{fuzzy}(q, e) = \alpha \cdot \text{TokenSet}(q, e) + (1{-}\alpha) \cdot \text{PartialRatio}(q, e)
\end{equation}
where $\alpha$ controls the trade-off between token-set overlap (handling reordered terms) and partial matching (handling substrings). Candidates below a threshold $\tau_{fuzzy}$ are discarded, yielding a high-precision entity set $E_{cat}$:
\begin{equation}
    E_{cat} = \{e \in C_{sem} : S_{fuzzy}(q, e) \geq \tau_{fuzzy}\}
\end{equation}

\subsubsection{Agentic External Search}
For long-tail or cold-start queries not yet indexed in the catalog, the model autonomously invokes an external web-search tool \cite{schick2023toolformer, qin2023toolllm, asai2024selfrag, yan2024corrective, azumo2025agentic} to retrieve real-time world knowledge $E_{ext}$ (Figure~\ref{fig:architecture}, step 6). This agentic capability allows the model to dynamically seek fresh context when internal confidence is low, helping it distinguish, for example, whether ``450 North'' is a craft brewery or a restaurant. Strategic business rules (e.g., a ``Restaurant-First Hierarchy'') are injected via a dynamic example bank \cite{brown2020gpt3}, enabling rapid policy updates without architectural changes.

\subsection{Dual-Intent Prediction and Disambiguation}
The grounded prompt, containing the query, retrieved entities $E_{cat}$, and any external signals $E_{ext}$, is passed to the LLM reasoning engine (Figure~\ref{fig:architecture}, step 5), which emits an ordered dual-intent tuple:
\begin{equation}
    (v_p, v_s) = \text{LLM}(q, E_{cat}, E_{ext}, \mathcal{P})
\end{equation}
where $v_p$ is the primary intent, $v_s$ is an optional secondary intent (or $\varnothing$), and $\mathcal{P}$ denotes the policy context (strategic overrides and few-shot examples).

When the model predicts a dual intent ($v_s \neq \varnothing$), the output is routed to a disambiguation layer (Figure~\ref{fig:architecture}, step 7). In this work, we implement a Pairwise Override function $\delta$ based on global historical win rates:
\begin{equation}
    v^* = \delta(v_p, v_s) = \begin{cases} v_s & \text{if } (v_p, v_s) \in \mathcal{W} \\ v_p & \text{otherwise} \end{cases}
\end{equation}
where $\mathcal{W}$ is a whitelist of conflict pairs for which historical data indicates the secondary intent should take priority. This layer is designed to be modular: it can be extended to incorporate personalization signals such as user location or order history, or replaced entirely with a learned re-ranker, without modifying the upstream prediction architecture.

\section{Experiments}
\label{sec: 4experiments}
\begin{table*}[t]
  \caption{Dual-intent resolution on ambiguous tail queries. Baseline: ungrounded Gemini-2.5-flash.}
  \label{tab:qualitative}
  \footnotesize
  \setlength{\tabcolsep}{4pt}
  \begin{tabular*}{\textwidth}{@{\extracolsep{\fill}}llllll}
    \toprule
    \textbf{Query} & \textbf{Baseline} & \textbf{Entity Context} & \textbf{Search Context} & \textbf{DI Output (p/s)} & \textbf{Resolved} \\
    \midrule
    ``wildflower'' & \textcolor{red}{Flower} & Wildflower Bites (Restaurant) & Popular restaurant chain in AZ & Restaurant / Flower & \textcolor{green!50!black}{\textbf{Restaurant}} \\
    ``better chew'' & \textcolor{red}{Pet Product} & Better Chew Farms (Plant-Based) & Plant-based frozen food brand & Grocery / Dish & \textcolor{green!50!black}{\textbf{Grocery}} \\
    ``450 north'' & \textcolor{red}{Restaurant} & 450 North Craft Ales (Beer) & Craft brewery, not a restaurant & Alcohol / Retail Store & \textcolor{green!50!black}{\textbf{Alcohol}} \\
    \bottomrule
  \end{tabular*}
\end{table*}

%\subsection{Setup}
We evaluate across four benchmarks representing distinct traffic slices ($N \approx 30{,}000$):
\textbf{Branded*} ($N{=}7{,}809$): retail brand-name queries;
\textbf{Retail} ($N{=}7{,}988$): general non-food retail queries;
\textbf{Tail (Synthetic)} ($N{=}2{,}335$): rare long-tail queries;
and \textbf{SOT} ($N{=}12{,}651$): a representative sample of global traffic across Head, Torso, and Tail segments. We compare against four baselines:
(1)~\textit{Legacy}: the existing production system, a hybrid BERT + LLM ensemble;
(2)~\textit{Gemini Base}: Gemini-2.5-flash with a standard single-intent prompt (ungrounded), serving as our control to isolate architectural contributions using the same backbone;
(3)~\textit{GPT-4o} and (4)~\textit{GPT-4o-mini}: ungrounded single-intent baselines representing a larger and a smaller general-purpose model.

\subsection{Results}
We report accuracy based on the final resolved intent after the disambiguation layer, i.e., the ground truth must match the single intent selected by our disambiguation policy from the predicted tuple $(v_p, v_s)$.

\begin{table}[t]
  \caption{Accuracy Comparison across Benchmarks.}
  \label{tab:results}
  \small
  \setlength{\tabcolsep}{0pt}
  \begin{tabular*}{\columnwidth}{@{\extracolsep{\fill}}lccccc}
    \toprule
    Benchmark & \shortstack{GPT\\4o\\Mini} & \shortstack{GPT\\4o} & \shortstack{Gemini\\Base} & Legacy & \shortstack{DI\\(Ours)}\\
    \midrule
    Branded* & 77.9\% & 87.5\% & 88.4\% & 82.3\% & \textbf{94.5\%} \\
    Retail & 86.5\% & 93.6\% & 90.5\% & 91.2\% & \textbf{96.3\%} \\
    Tail (Synthetic) & 82.6\% & 87.1\% & 86.4\% & 83.4\% & \textbf{92.4\%} \\
    SOT (Overall) & 79.8\% & 83.2\% & 83.1\% & 89.4\% & \textbf{94.0\%} \\
    \bottomrule
  \end{tabular*}
\end{table}

The system achieves material accuracy gains across all benchmarks (Table~\ref{tab:results}). On the global SOT benchmark, it outperforms the Gemini baseline by +10.9pp, GPT-4o by +10.8pp, and GPT-4o-mini by +14.2pp, confirming that general-purpose LLM reasoning is prone to systematic intent misalignment when disconnected from platform-specific inventory data. The gap widens further on the Tail segment (+13.6pp vs GPT-4o, +16.1pp vs GPT-4o-mini).

\begin{figure}[ht]
  \centering
  \begin{tikzpicture}
  \begin{axis}[
      ybar,
      enlarge x limits=0.25,
      legend style={at={(0.5,-0.2)},
        anchor=north,legend columns=-1},
      ylabel={Accuracy (\%)},
      symbolic x coords={Head, Torso, Tail},
      xtick=data,
      nodes near coords,
      every node near coord/.append style={font=\scriptsize},
      nodes near coords align={vertical},
      ymin=60, ymax=105,
      width=\columnwidth,
      height=6cm,
      bar width=10pt,
      cycle list={
          {fill=gray!50,draw=black},     % GPT-4o-mini
          {fill=brown!30,draw=brown},    % GPT-4o
          {fill=red!30,draw=red},        % Gemini Base
          {fill=blue!30,draw=blue},      % Legacy
          {fill=green!60!black,draw=black} % DI (Ours) - GREEN
      }
  ]
  \addplot coordinates {(Head,88.7) (Torso,80.6) (Tail,74.6)}; % GPT-4o-mini
  \addplot coordinates {(Head,93.1) (Torso,84.4) (Tail,77.1)}; % GPT-4o
  \addplot coordinates {(Head,92.3) (Torso,83.9) (Tail,77.7)}; % Baseline
  \addplot coordinates {(Head,98.4) (Torso,92.2) (Tail,82.2)}; % Cache
  \addplot coordinates {(Head,99.5) (Torso,94.7) (Tail,90.7)}; % DI
  \legend{GPT-4o-mini, GPT-4o, Gemini Base, Legacy, DI (Ours)}
  \end{axis}
  \end{tikzpicture}
  \caption{SOT Accuracy by Segment.}
  \Description{Bar chart comparing GPT-4o-mini, GPT-4o, Gemini Base, the legacy system, and the proposed dual-intent system across Head, Torso, and Tail traffic. The dual-intent system is best on all segments, with the largest gap on Tail queries where it reaches 90.7 percent versus 82.2 percent for the legacy system and 77.7 percent for Gemini Base.}
  \label{fig:sot}
  \end{figure}
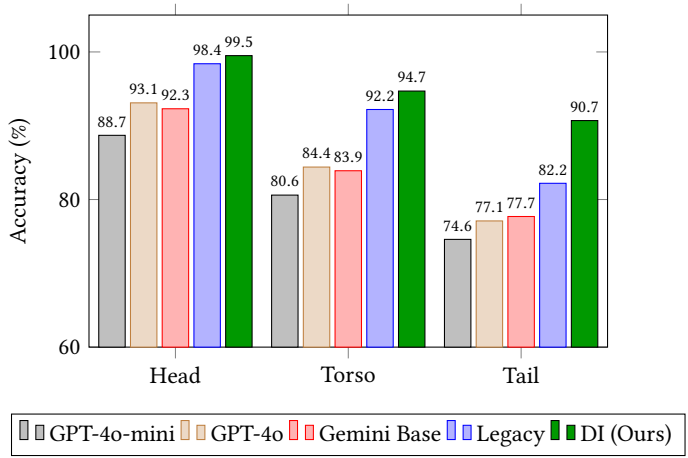

To understand where the system provides the most value, we performed a segmented analysis of the SOT benchmark (Figure~\ref{fig:sot}). While the Gemini baseline struggles on the Tail (77.7\%) and Torso (83.9\%) due to data sparsity, our system achieves 90.7\% on the Tail, surpassing the legacy system by +8.5pp. The system remains safe on high-volume Head queries (${\sim}$99\% accuracy), confirming that the multi-intent framework improves long-tail coverage without compromising navigational reliability.

\subsection{Ablation and Deployment}
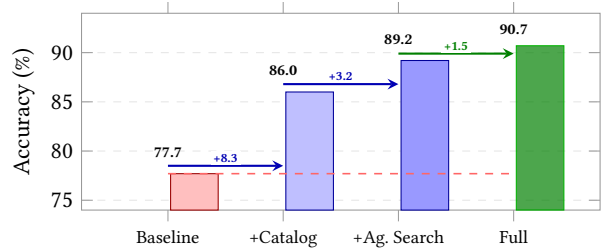
\begin{figure}[ht]
\centering
\begin{tikzpicture}
\begin{axis}[
    ybar,
    ylabel={Accuracy (\%)},
    symbolic x coords={Baseline, +Catalog, +Ag. Search, Full},
    xtick=data,
    x tick label style={font=\footnotesize},
    ymin=74, ymax=94,
    width=\columnwidth,
    height=4.2cm,
    bar width=18pt,
    enlarge x limits=0.25,
    ytick={75,80,85,90,95},
    ymajorgrids=true,
    grid style={dashed, gray!20},
    axis line style={gray!80},
    tick style={gray!80},
]
\addplot[fill=red!25, draw=red!60!black, forget plot] coordinates {(Baseline, 77.7)};
\addplot[fill=blue!25, draw=blue!60!black, forget plot] coordinates {(+Catalog, 86.0)};
\addplot[fill=blue!40, draw=blue!60!black, forget plot] coordinates {(+Ag. Search, 89.2)};
\addplot[fill=green!50!black, draw=green!70!black, fill opacity=0.7, forget plot] coordinates {(Full, 90.7)};
\node[above=3pt, font=\scriptsize\bfseries] at (axis cs:Baseline, 77.7) {77.7};
\node[above=3pt, font=\scriptsize\bfseries] at (axis cs:+Catalog, 86.0) {86.0};
\node[above=3pt, font=\scriptsize\bfseries] at (axis cs:+Ag. Search, 89.2) {89.2};
\node[above=1pt, font=\scriptsize\bfseries] at (axis cs:Full, 90.7) {90.7};
\draw[dashed, red!60, line width=0.6pt] (axis cs:Baseline, 77.7) -- (axis cs:Full, 77.7);
\draw[-stealth, thick, blue!70!black] (axis cs:Baseline, 78.5) -- node[above, font=\tiny\bfseries, fill=white, inner sep=1pt] {+8.3} (axis cs:+Catalog, 78.5);
\draw[-stealth, thick, blue!70!black] (axis cs:+Catalog, 86.8) -- node[above, font=\tiny\bfseries, fill=white, inner sep=1pt] {+3.2} (axis cs:+Ag. Search, 86.8);
\draw[-stealth, thick, green!50!black] (axis cs:+Ag. Search, 89.9) -- node[above, font=\tiny\bfseries, fill=white, inner sep=1pt] {+1.5} (axis cs:Full, 89.9);
\end{axis}
\end{tikzpicture}
\caption{Component Ablation on SOT Tail ($N{=}4{,}993$). Cumulative +13.0pp lift over the ungrounded baseline.}
\Description{Bar chart showing incremental gains from each component on the SOT Tail benchmark. Accuracy rises from 77.7 percent for the ungrounded baseline to 86.0 percent with catalog grounding, 89.2 percent after adding agentic search, and 90.7 percent for the full system with dual-intent disambiguation.}
\label{fig:ablation}
\end{figure}

To quantify the contribution of each architectural component, we conduct an incremental ablation on the SOT Tail ($N{=}4{,}993$), where intent ambiguity is most prevalent. 
Starting from the ungrounded baseline, we progressively add Catalog Grounding, then Agentic Search, and finally Dual-Intent Disambiguation.
As shown in Figure~\ref{fig:ablation}, Catalog Grounding provides an appreciable lift (+8.3pp), confirming that proprietary inventory context is the key driver of accuracy. Adding Agentic Search yields a further +3.2pp by grounding the model on fresh, real-time web context for cold-start and ambiguous queries that may be absent from the catalog. Dual-Intent Disambiguation contributes +1.5pp by capturing secondary interpretations that a single-label system discards. Together, these yield a cumulative +13.0pp lift over baseline. Table~\ref{tab:qualitative} illustrates how these components resolve real ambiguous queries.

The system is deployed in production via an offline batch-and-cache architecture, currently covering 95.9\% of daily search. Cache misses fall back to an online BERT-based classifier, preserving full coverage.
Preliminary online results align with the offline gains: in the changed/disagreement segment, treatment improves CTR from 31.74\% to 34.36\% and query purchase rate (QPR) from 6.66\% to 6.90\%.
The offline teacher also produces a silver-standard dataset for future distilled student models that can replace the BERT fallback with a lower-latency model that mimics the grounded teacher \cite{hinton2015distilling, srinivasan-etal-2022-quill, boateng-etal-2025-concept, boateng2025weaklm}.

\section{Conclusion and Future Work}
\label{sec: 5conclusion}
We presented an Agentic Multi-Source Grounded system for query intent understanding that combines catalog grounding, agentic web search, and dual-intent prediction with pluggable disambiguation. Evaluated on DoorDash's multi-category search platform, the system achieves +13.0pp accuracy on long-tail queries over the ungrounded baseline.
These results show that agentic multi-source grounding is a practical production strategy for resolving ambiguous marketplace queries at scale, and motivate future distilled student models for lower-latency online serving.
A future direction is to distill the grounded offline teacher into lower-latency student models, extending ambiguity-resolution gains beyond cached queries and enabling broader online deployment.

\begin{acks}
We thank Elyse Winer, Steven Xu, Raghav Saboo, and Guangxiang Chen for infrastructure, productionization, and product and technical support that informed this work.
\end{acks}

\bibliographystyle{ACM-Reference-Format}
\balance
\bibliography{references}
\end{document}